\documentclass[]{spie} 

\usepackage{amsmath,amsfonts,amssymb}
\usepackage{graphicx}
\usepackage{wrapfig}
\usepackage[colorlinks=true, allcolors=blue]{hyperref}

\usepackage[long]{longShort}

\usepackage{booktabs}

\title{Evaluating the Impact of Intensity Normalization\\on MR Image Synthesis}

\author[a]{Jacob C. Reinhold}
\author[a,b]{Blake E. Dewey}
\author[a,c]{Aaron Carass}
\author[a,c]{Jerry L. Prince}
\affil[a]{Department of Electrical and Computer Engineering, Johns Hopkins
University, Baltimore,~MD,~USA~21218}
\affil[b]{F.M. Kirby Center for Functional Brain Imaging, Kennedy Krieger
Institute, Baltimore,~MD,~USA,~21205}
\affil[c]{Department of Computer Science, Johns Hopkins University,
Baltimore,~MD,~USA~21218}

\authorinfo{Further author information: (Send correspondence to J.R.)\\J.R.: E-mail: jacob.reinhold@jhu.edu}

\begin{document} 
\maketitle
\begin{abstract}

  Image synthesis learns a transformation from the intensity features
  of an input image to yield a different tissue contrast of the output
  image. This process has been shown to have application in many
  medical image analysis tasks including imputation, registration, and
  segmentation.  To carry out synthesis, the intensities of the input
  images are typically scaled---i.e., normalized---both in training to
  learn the transformation and in testing when applying the
  transformation, but it is not presently known what type of input
  scaling is optimal.  In this paper, we consider seven different intensity
  normalization algorithms and three different synthesis methods to
  evaluate the impact of normalization. Our experiments demonstrate
  that intensity normalization as a preprocessing step improves the
  synthesis results across all investigated synthesis
  algorithms. Furthermore, we show evidence that suggests intensity
  normalization is vital for successful deep learning-based MR image
  synthesis.

\end{abstract}


\section{Introduction}
\label{sec:intro}

For magnetic resonance (MR) images, we can view image synthesis as
learning an intensity transformation between two differing contrast
images, e.g., from T1-weighted~(T1-w) to T2-weighted (T2-w) or FLuid
Attenuated Inversion Recovery (FLAIR). Synthesis can
generate contrasts not present in the data set---i.e., image
imputation---which are useful for 
image processing applications such as registration and
segmentation~\cite{Iglesias2013,huo2017adversarial}. The transformation need not be limited to
MR images; an example application is MR to computed tomography~(CT)
registration where it has been shown to improve accuracy when the
moving image is synthesized to match the target image's
contrast~\cite{Roy2014}. \Long{Other examples include multi-contrast
skull-stripping for MR brain images~\cite{Roy2017}, which performs
better with synthesized T2-w images when the original T2-w images
are unavailable.}

Methods to carry out image synthesis include sparse
recovery-based methods~\cite{Roy2013}, random forest
regression~\cite{Jog2017,Zhao2017}, registration~\cite{Lee2017,Cardoso2015}, and deep
learning~\cite{Wolterink2017,Chartsias2018}. Evidence suggests that 
accurate synthesis is heavily dependent on a standard intensity scale across
the sample of images used in the training procedure. That is \Long{to
successfully train a synthesis algorithm} the training and testing
data must have similar intensity properties\Long{ (e.g., the mean
  intensity of white matter should be the same for all input
  images)}. This is a problem in MR synthesis since MR
images do not have a standard intensity scale.

\Short{\section{New Work to be Presented}}

In this paper, we explore seven methods to normalize the intensity
distribution of a sample of MR brain images within each of three
contrasts (T1-w, T2-w, and FLAIR). We then quantitatively compare
their performance in the task of synthesizing T2-w and FLAIR images
from T1-w contrasts using three synthesis algorithms. We show results
that suggest intensity normalization as a preprocessing step is
crucial for consistent MR image synthesis. \Short{In this abstract, we
only present T1-w to FLAIR synthesis results over six subjects. In
the final paper we will include both T1-w to FLAIR and T1-w to T2-w
synthesis over a cohort of 20 subjects. We also provide a terse
summary of the normalization techniques here and will provide
detailed descriptions in the final paper.}

\section{METHODS}

\Short{We first outline the seven intensity normalization algorithms
  considered in this paper, then describe the three different
  synthesis methods. We denote an MR brain image $I(\mathbf x)$ with
  $\mathbf x \in [0,N] \times [0,M] \times [0,L] \subset \mathbb{N}^3$
  for $N, M, L \in \mathbb{N}$. Let $W$ and $B$, denote masks for the
  white matter~(WM) and brain mask, respectively, with $W \subset B
  \subset I$. ($W$ is segmentation dependent and described below.)
  $\mu_{\cdot}$ and $\sigma_{\cdot}$ are the mean and standard
  deviation, respectively, of the intensities within $\cdot$ where
  $\cdot$ is one of $W$, $B$, or $I$.}

\Long{

In this section, we first describe the seven intensity normalization
algorithms considered in this paper, namely: 1)~Z-score, 2)~Fuzzy
C-Means (FCM)-based, 3)~Gaussian mixture model (GMM) based, 4)~Kernel
Density Estimate (KDE) based, 5)~Piecewise linear histogram matching
(HM)~\cite{Nyul2000, Shah2011}, 6)~WhiteStripe~\cite{Shinohara2014},
and 7)~RAVEL~\cite{Fortin2016}. We then describe three different synthesis
routines: 1)~polynomial regression, 2)~random forest regression, and
3)~deep neural network based synthesis. For the following
subsections, let $I(\mathbf x)$ be the MR brain image under
consideration where $\mathbf x \in [0,N]\times[0,M]\times[0,L]\subset
\mathbb{N}^3$ for $N,M,L \in \mathbb{N}$, the dimensions of $I$, and
let $B \subset I$ be the corresponding brain mask (i.e., the set of
indices corresponding to the location of the brain in $I$).

}

\subsection{Normalization}

\Short{
In Table~\ref{t:norm}, we provide an overview of the 
normalization techniques included in this paper. For segmentation
based normalization we use three different segmentation algorithms:
1)~three class fuzzy c-means~($I_{\text{\tiny{FCM}}}$), 2)~three class
Gaussian mixture model~($I_{\text{\tiny{GMM}}}$), and 3)~a kernel
density estimate~($I_{\text{\tiny{KDE}}}$) based approach. Each of
which provides a definition of WM which is used to define $W$ and
compute $\mu_W$ and $\sigma_W$ in the normalization formulation.
WhiteStripe~\cite{Shinohara2014} uses WM values within 5\% of the WM
peak with respect to the cumulative distribution function~(CDF). RAVEL
normalization~\cite{Fortin2016} builds upon WhiteStripe by removing
unwanted technical variation from a sample of images with a voxel
specific correction. RAVEL requires registration to a normalized
space, for which we used SyN~\cite{Avants2008}.

\begin{table}[!tb]
\centering
\caption{This is a list of the proposed normalization methods. The
segmentation based normalization has three varieties that are detailed
in the paper.}
\label{t:norm}
\begin{tabular}{l c l}
\toprule
\textbf{Method} && \textbf{Formulation}\\
\cmidrule{1-1}
\cmidrule{3-3}
Z-score && $\displaystyle{I_{\text{\tiny{z-score}}}(\mathbf x) =
\frac{ I(\mathbf x) - \mu_B }{ \sigma_B }}$\\[0.4em]
Segmentation based && $\displaystyle{I_{\text{\tiny{SEG}}}(\mathbf x)
= c\frac{ I(\mathbf x) }{ \mu_W }}$ \quad where $c = 1000$\\[0.4em]
PLHM && \begin{minipage}{0.7\textwidth}Piecewise linear histogram
matching~\cite{Nyul2000, Shah2011} based on deciles.\end{minipage}\\
WhiteStripe && $\displaystyle{I_{\text{\tiny{WS}}}(\mathbf x) = \frac{
I(\mathbf x) - \mu_{\text{\tiny{WS}}} }{ \sigma_{\text{\tiny{WS}}}} }$
\quad where $\text{WS} = \pm5\%$ of the WM CDF\\[0.4em]
RAVEL && \begin{minipage}{0.7\textwidth}$\displaystyle{ I_{
\text{\tiny{RAVEL}} }(\mathbf x) = I_{ \text{\tiny{WS}}} -
\gamma_{\mathbf x} }$ where $\gamma_{\mathbf x}$ is a voxel dependent
correction factor for unwanted variation.\end{minipage}\\
\bottomrule
\end{tabular}
\end{table}
}

\Long{

  In the following sections, we will briefly overview the intensity
  normalization algorithms used in this experiment. Code for the
  following intensity normalization algorithms is at:
  \url{https://github.com/jcreinhold/intensity-normalization}.

\subsubsection{Z-score}

Z-score normalization uses the brain mask $B$ for the image $I$ to
determine the mean $\mu_{\text{zs}}$ and standard deviation $\sigma_{\text{zs}}$ of the
intensities inside the brain mask. Then the Z-score normalized image
is
\[ I_{\text{z-score}}(\mathbf x) = \frac{I(\mathbf x) - \mu_{\text{zs}}}{\sigma_{\text{zs}}}. \]

\subsubsection{FCM-based}

FCM-based normalization uses fuzzy c-means to calculate a white matter (WM) mask of the image $I$. 
This WM mask is then used to normalize the entire image to the mean of the WM. The procedure is as follows. 
Let $W\subset B$ be the WM mask for the image $I$, i.e., $W$ is the set of indices 
corresponding to the location of the WM in the image $I$. Then the WM mean is 
$\mu_{\text{fcm}} = \frac{1}{|W|} \sum_{\mathbf w \in W} I(\mathbf w)$.
and the FCM-based normalized image is
\[ I_{\text{fcm}}(\mathbf x) = \frac{c_1\cdot I(\mathbf x)}{\mu_{\text{fcm}}}, \]
where $c_1 \in \mathbb{R}_{>0}$ is a constant that determines the WM mean after normalization. 
In this experiment, we use three-class fuzzy c-means to get a segmentation of the WM over the brain mask
$B$ for the T1-w image and we arbitrarily set $c_1 = 1000$.

\subsubsection{GMM-based}
GMM-based normalization fits a mixture of three normal distributions to the
histogram of intensities inside the brain mask. The mean $\mu_{\text{gmm}}$ of the mixture
component associated with the WM is then used in the same way as the
FCM-based method, so the GMM-based normalized image is
\[ I_{\text{gmm}}(\mathbf x) = \frac{c_2\cdot I(\mathbf x)}{\mu_{\text{gmm}}}, \]
where $c_2 = 1000$ is a constant that determines the WM mean after normalization. 
The WM mean $\mu_{\text{gmm}}$ is determined by picking the
mixture component with the maximum intensity mean for T1-w images, the middle
intensity mean for FLAIR images, and the minimum intensity mean for T2-w images.

\subsubsection{Kernel Density Estimate-based}

KDE-based normalization estimates the empirical probability density function (pdf) of the
intensities of $I$ over the brain mask $B$ using the method of kernel density
estimation. In our experiment, we use a Gaussian kernel. The kernel density estimate 
provides a smooth version of the histogram
which allows us to more robustly pick the maxima associated with the WM via a
peak finding algorithm. The found WM peak $\rho$ is then used to
normalize the entire image, in much the same way as FCM-based
normalization. Namely,
\[ I_{\text{kde}}(\mathbf x) = \frac{c_3 \cdot I(\mathbf x)}{\rho}, \]
where $c_3 = 1000$ is a constant that determines the WM peak after normaalization. 
The WM peak is determined in T1-w and FLAIR by picking the
peak associated with the greatest intensity (for FLAIR, this is due to the
inability to distinguish between the WM and GM peaks) and for T2-w images the WM
peak is determined by the highest peak.

\subsubsection{Piecewise Linear Histogram Matching}

Piecewise linear histogram matching\cite{Nyul2000} (which we denote as HM for
brevity) addresses the normalization problem by learning a standard histogram
for a set of contrast images and linearly mapping the intensities of each
image to this standard histogram. The standard histogram is learned through
averaging pre-defined landmarks of interest on the histogram of a set of images. 
In Shah et al.\cite{Shah2011}, the authors demonstrate good results with
this method by defining landmarks as intensity percentiles at
$1,10,20,\ldots,90,99$ percent (where the intensity values below 1\% and above
99\% are extrapolated from the [1,10] and [90,99] percent intervals). We use these 
landmarks in our method and arbitrarily set the range of the standard scale to $[1,100]$. 
The intensity values of the set of images are then mapped piecewise linearly to the 
learned standard histogram along the landmarks. For further detail into the method
see Ny\'ul et al.\cite{Nyul2000} and Shah et al.\cite{Shah2011}.

\subsubsection{WhiteStripe}

WhiteStripe intensity normalization \cite{Shinohara2014} performs a
Z-score normalization based on the intensity values of normal appearing white
matter (NAWM). The NAWM is found by smoothing the histogram of the image and
selecting the highest intensity peak for T1-w images (the peaks for the other
contrasts are determined in the same way as described in the KDE section). 
Let $\mu_{\text{ws}}$ be the intensity associated with this peak. The ``white stripe'' is
then defined as the 10\%
segment of intensity values around $\mu_{\text{ws}}$. That is, let $F(x)$ be the cdf of the
specific MR image $I(\mathbf x)$ inside its brain mask $B$, and define $\tau =
5\%$. Then, the white stripe $\Omega_\tau$ is defined as the set
\[ \Omega_\tau = \left\{I(\mathbf x) \mid F^{-1}\left(F(\mu_{\text{ws}}) - \tau\right) < I(\mathbf
x) < F^{-1}\left(F(\mu_{\text{ws}}) + \tau\right)\right\}. \]
Let $\sigma_{\text{ws}}$ be the sample standard deviation associated with $\Omega_\tau$.
Then the WhiteStripe normalized image is
\[ I_{\text{ws}}(\mathbf x) = \frac{I(\mathbf x) - \mu_{\text{ws}}}{\sigma_{\text{ws}}}. \]

\subsubsection{RAVEL}

RAVEL normalization \cite{Fortin2016} adds an additional normalization step to WhiteStripe by
removing unwanted technical variation (defined below) from a sample of $m$ images. Following 
the notation in the original paper\cite{Fortin2016}, the method assumes
that cerebrospinal fluid (CSF) is associated with technical variation, and---after WhiteStripe
normalization---the CSF intensities can be written as
\[ V_c = \gamma Z^\top + R, \]
where $V_c$ is an $n \times m$ matrix of CSF intensities, $\gamma Z^\top$ represents the unknown
technical variation, and $R$ is a matrix of the residuals. The $n$ CSF intensity
values in $V_c$ are determined by deformably co-registering the images,
finding a CSF mask for each deformably registered image, and taking the
intersection across all the masks.

We then use singular value decomposition to write $V_c = U \Sigma W^\top$. 
Then $W$ is an $m \times m$ matrix of right singular vectors
and we can use $b\le m$ right singular vectors to form a
linear basis for the unwanted factors $Z$\cite{Leek2007}, where $b$ is the unknown true
rank of $V_c$. That is, we use $W_b$ as a surrogate for 
$Z$, where $W_b$ is the subset of $b$ columns of $W$ collected into a matrix. We then do
voxel-wise linear regression to estimate the coefficients $\gamma$. The
RAVEL normalized image is then defined as
\[ I_{\text{ravel}}(\mathbf x) = I_{\text{ws}}(\mathbf x) - \gamma_{\mathbf x} W_b^\top, \]
where $\gamma_{\mathbf x}$ are the coefficients of unwanted variation associated
with the voxel $\mathbf x$ found through linear regression. In our experiments, 
we follow the original paper\cite{Fortin2016} and fix $b=1$\footnote{The first right singular vector is highly correlated (>95\%)
with the mean intensity of the CSF\cite{Fortin2016}.}. For deformable registration, we use
SyN\cite{Avants2008} to register all images to one image in the data set.

}

\subsection{Synthesis}
\Short{Image synthesis can be thought of as a regression on image
intensities, i.e., learning a parametric or non-parametric mapping
from one contrasts intensity distribution to another contrasts
intensity distribution. We explore three image synthesis methods:
1)~polynomial regression~(PR), 2)~random forest~(RF) regression, and
3)~deep neural network~(DNN)-based synthesis.

PR samples 100,000 voxels in $B$ and train a regressor from the
intensities of $\mathbf{x}$ and its six connected neighbors in the
source image to the intensity of $\mathbf{x}$ in the target image. We
use a third-order polynomial regressor to learn the mapping from the
source intensity to the target intensity. Our RF regression is based
on a simplified model of the work of Jog~et~al.~\cite{Jog2017}.
Training is done from 100,000 sampled voxels with the image patch
using voxels in the six primary directions at 3, 5, and 7 indices away
from the center voxel and no multi-scale framework. This simplified RF
model also only synthesizes the corresponding center voxel of the
patch under consideration.

Our DNN approach uses a 4-level U-net~\cite{ronneberger2015miccai},
extracting $128 \times 128$ patches from axial, sagittal, and coronal
orientations to learn the synthesis, in a similar architecture to
Zhao~et~al.~\cite{Zhao2017a}. We use instance normalization and leaky
ReLUs with parameter 0.2 as the activation function since Z-score,
WhiteStripe, and RAVEL allow for negative values in the images. It is
a good approximation for the state-of-the-art in image synthesis.
Complete detail of all three synthesis approaches will be included in
the final paper.}

\Long{
Image synthesis can be described as a regression on the intensities of the
images, i.e., learning a parametric or non-parametric mapping from one
contrasts intensity distribution to another contrasts intensity
distribution. In this section we describe three methods of image synthesis: 
1) polynomial regression (PR), 2) random forest regression (RF), and 3) deep neural 
network (DNN)-based synthesis. 

\subsubsection{Polynomial Regression}

For polynomial regression, we randomly select 100,000 voxels
inside the brain mask. For the source images, we extracted 
patches around each of these voxels where the patches include the center voxel 
and its six neighbors. For the target images, we extract only the corresponding center voxel. We extract the patches in this way across
all images, so for $M$ images we have an $(M \cdot 100,000) \times 7$ feature matrix for
the source images and an $(M \cdot 100,000) \times 1$ feature matrix for the target
images. We use a third-order polynomial as the regressor to learn the mapping
from the source feature matrix to the target feature matrix. We use
this na\"ive model to provide a low-variance baseline for image synthesis methods.

\subsubsection{Random Forest Regression}

Similar to polynomial regression, in random forest regression---inspired
by Jog, et al.\cite{Jog2017}---we randomly select 100,000 voxels
inside the brain mask. For the source images, we extracted 
patches that comprise the center voxel, its six neighbors, and the voxels in the six primary
directions at 3, 5, and 7 voxels away from the center. For the target images, we
extract only the corresponding center voxel. We extract the patches in this way across
all images, so for $M$ images we have an $(M \cdot 100,000) \times 25$ feature matrix for
the source images and an $(M \cdot 100,000) \times 1$ feature matrix for the target
images. For the random forest regressor that learns the mapping between the
source feature matrix and the target feature matrix, we set the number of trees to 60 and
the number of samples in a leaf node to 5.

\subsubsection{DNN}

We use a 4-level U-net\cite{Ronneberger2015} and extract $128 \times 128$
patches from axial, sagittal, and coronal orientations to learn the synthesis.
Patches are extracted in this fashion for data augmentation.
We use instance normalization and leaky ReLUs with parameter 0.2 as the
activation function since Z-score, WhiteStripe, and RAVEL allow for negative values in the images. The
architecture follows Zhao, et al.\cite{Zhao2017a} who used a similar structure for a synthesis
task. We trained the network for 100 epochs for all sets of normalized images
excluding the unnormalized images with which we trained the DNN for 400 epochs. This
discrepancy in the number of epochs used is due to a failure of convergence observed 
in the first 100 epochs for the unnormalized images.
}

\subsection{Quality Assessment}
We use three different metrics to quantitatively determine the
performance of the synthesis result. Note that all three metrics
compare the result to the ground truth images which were not used in training any 
synthesis methods. The metrics are:
1)~normalized cross-correlation (NCC), 2)~mean structural
similarity~(MSSIM)~\cite{Wang2004}, and 3)~mutual information~(MI). We
use these metrics as opposed to MSE or PSNR as the data have been
scaled to different ranges, making MSE and PSNR not easily comparable
across normalization routines.

\section{RESULTS}
For evaluation, we use 18 data sets from the Kirby-21
data set\Long{~\cite{Landman2011}}. All of the subjects for the data sets are
verified to be healthy subjects. From
these 18 data sets, we use the T1-w, T2-w, and FLAIR images. All the
images are resampled to 1mm$^3$, bias field corrected with
N4\Long{~\cite{Tustison2010}}, and each T2-w and FLAIR image is
affinely registered to the corresponding T1-w image with the ANTs
package\Long{~\cite{avants2009advanced}}. The brain mask for the
images are found with ROBEX\Long{~\cite{Iglesias2011}} and the mask is
used during normalization and applied to the images before synthesis
such that the background is zero in all the images.

\begin{figure}[!ht]
\centering
\includegraphics[width=\textwidth]{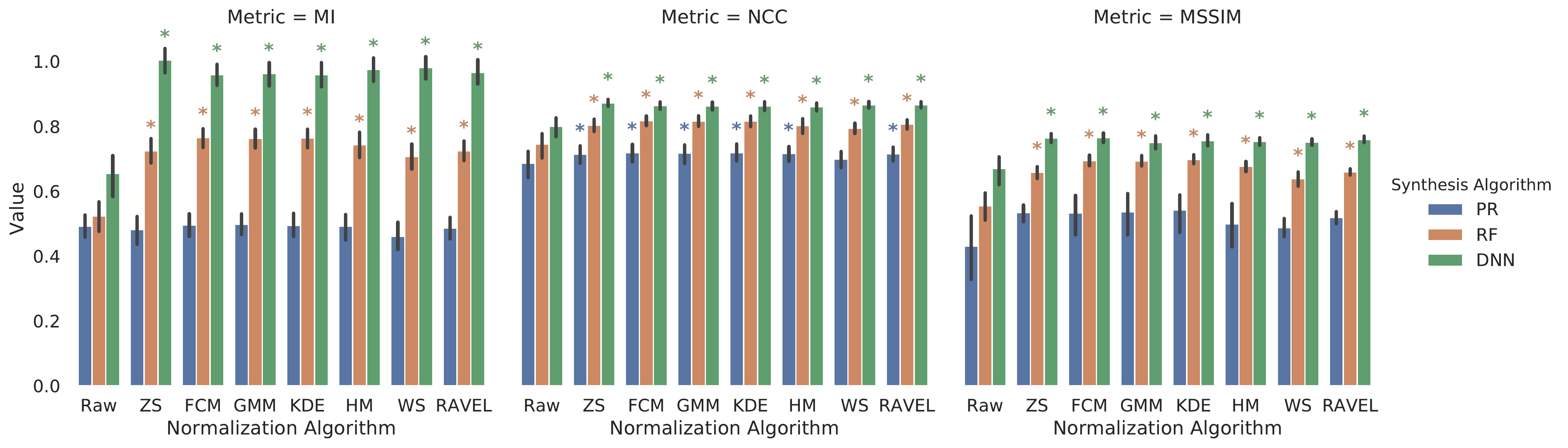}
\caption{\label{fig:barchartflair}\textbf{T1-to-FLAIR Quality
Metrics:} Raw corresponds to synthesis using unnormalized images, ZS to Z-score normalized images, and WS to WhiteStripe normalized images.
Statistical significance (denoted by *) for each experiment is compared to Raw ($p < 0.05$). The error bars represent the 95\% confidence interval.}
\end{figure}

\begin{figure}[!tb]
\centering
   \begin{tabular}{cccc} 

   \textbf{PR} & \textbf{RF} & \textbf{DNN} & \textbf{Truth}\\

   \includegraphics[width = 0.20\textwidth, clip, trim = 1cm 1.3cm 1cm 1.3cm]{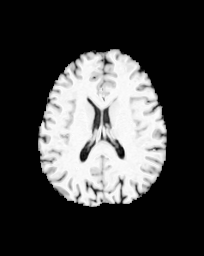} &
   \includegraphics[width = 0.20\textwidth, clip, trim = 1cm 1.3cm 1cm 1.3cm]{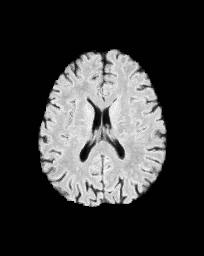} &
   \includegraphics[width = 0.20\textwidth, clip, trim = 1cm 1.3cm 1cm 1.3cm]{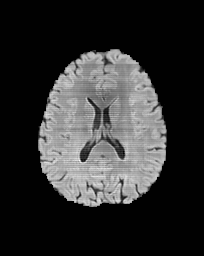} &
   \includegraphics[width = 0.20\textwidth, clip, trim = 1cm 1.3cm 1cm 1.3cm]{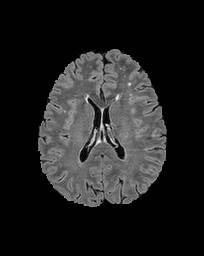}\\
   \includegraphics[width = 0.20\textwidth, clip, trim = 1cm 1.3cm 1cm 1.3cm]{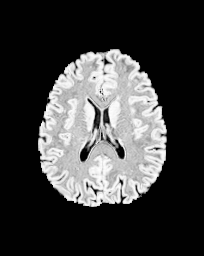} &
   \includegraphics[width = 0.20\textwidth, clip, trim = 1cm 1.3cm 1cm 1.3cm]{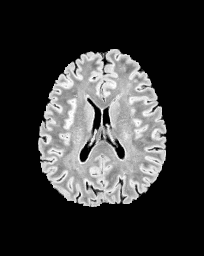} &
   \includegraphics[width = 0.20\textwidth, clip, trim = 1cm 1.3cm 1cm 1.3cm]{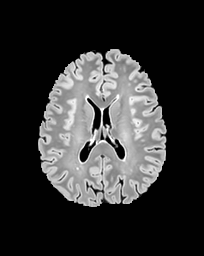} &
   \includegraphics[width = 0.20\textwidth, clip, trim = 1cm 1.3cm 1cm 1.3cm]{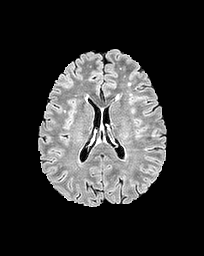} \\

   \end{tabular}
\caption{\label{fig:flairsynth}\textbf{T1-to-FLAIR Synthesis results:}
Shown are the results of synthesis using unnormalized~(top row) and
FCM normalized images~(bottom row). The unnormalized DNN result
represents a failure of image synthesis.}
\end{figure}

We split the data into two sets of nine for training and nine for testing. 
Bar charts in Figs.~\ref{fig:barchartflair} and~\ref{fig:barchartt2} show
the mean and the bootstrapped 95\% confidence interval of the T1-to-FLAIR and T1-to-T2 synthesis, 
respectively, for the quality metrics averaged over all testing data sets, for every
normalization scheme and synthesis algorithm. We use the Wilcoxon 
signed-rank test to compare the distributions of each normalized method, for all metrics,
against the corresponding unnormalized results per synthesis algorithm. We use a 
statistical significance level of $\alpha = 0.05$ and show that this threshold is met 
in Figs.~\ref{fig:barchartflair} and~\ref{fig:barchartt2} with an asterisk above the 
corresponding bar. Figures~\ref{fig:flairsynth} and~\ref{fig:t2synth} show results for the various synthesis
algorithms with unnormalized training data~(denoted raw) and
normalized training data use the FCM approach. 

\Long{
The experiments show that
synthesis results are robust to the choice of normalization algorithm,
which are stable around the same levels across all metrics. This qualitative
result, observed in Figs.~\ref{fig:barchartflair} and~\ref{fig:barchartt2}, is reinforced 
with statistical tests. We use the Wilcoxon
signed-rank test (with Bonferroni correction) to show statistically
significant difference between any of the presented normalization algorithms for each
metric ($\alpha = 0.05$); however, no normalization algorithm consistently met this threshold for 
any metric with any synthesis algorithm in either T1-to-FLAIR or T1-to-T2 synthesis.
An interesting finding is that---in T1-to-T2 synthesis---the random forest regressor 
qualitatively performs more robustly on unnormalized data, but both the DNN and polynomial
regression methods fail; in terms of NCC, the DNN synthesis has zero
mean because of negative correlation in some of the testing results. 

Failure cases of synthesis in T1-to-FLAIR and T1-to-T2
unnormalized images are shown in Figs.~\ref{fig:flairsynth} and~\ref{fig:t2synth},
respectively, which can be compared to the successfully
synthesized FCM-normalized images in the same figures. We discuss
in the following section.
}

\Long{
\begin{figure}[!ht]
\begin{center}
\includegraphics[width=\textwidth]{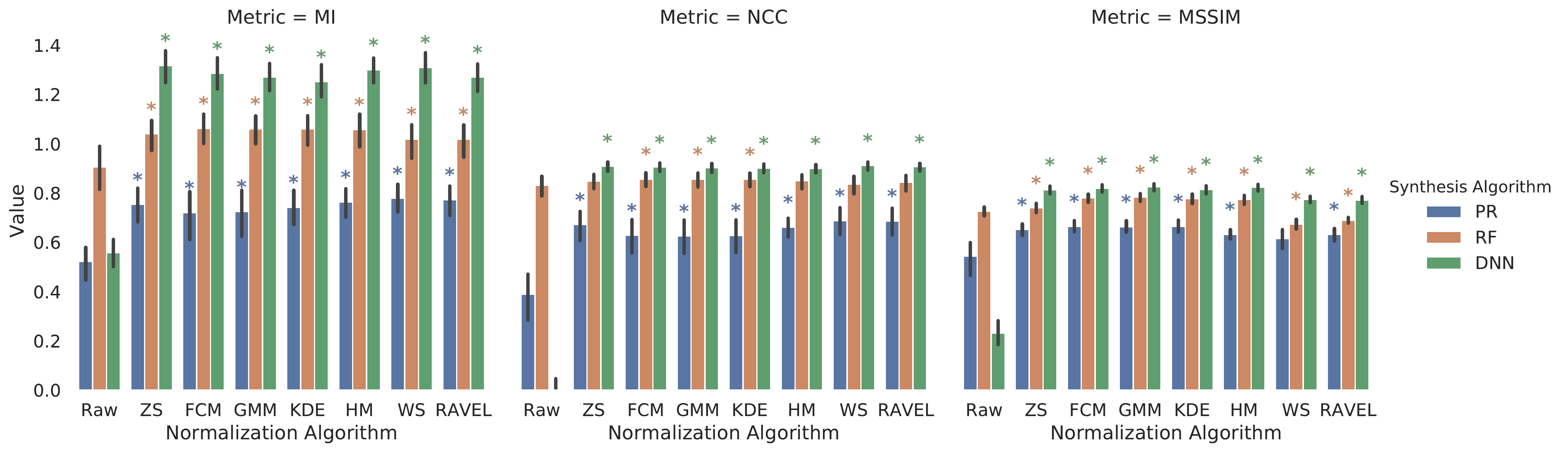}
\caption{\label{fig:barchartt2}\textbf{T1-to-T2 Quality
Metrics:} Raw corresponds to synthesis using unnormalized images, ZS to Z-score normalized images, and WS to WhiteStripe normalized images.
Statistical significance (denoted by *) for each experiment is compared to Raw ($p < 0.05$). The error bars represent the 95\% confidence interval.}
\end{center}
\end{figure}
}

\Long{
\begin{figure}[!ht]
\centering
   \begin{tabular}{cccc} 

   \textbf{PR} & \textbf{RF} & \textbf{DNN} & \textbf{Truth}\\

   \includegraphics[width = 0.20\textwidth, clip, trim = 1cm 1.3cm 1cm 1.3cm]{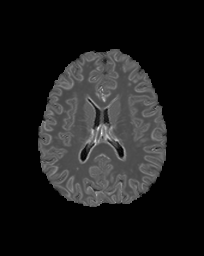} &
   \includegraphics[width = 0.20\textwidth, clip, trim = 1cm 1.3cm 1cm 1.3cm]{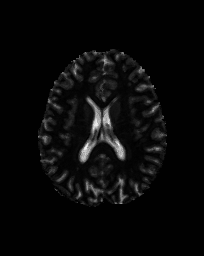} &
   \includegraphics[width = 0.20\textwidth, clip, trim = 1cm 1.3cm 1cm 1.3cm]{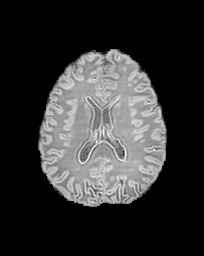} &
   \includegraphics[width = 0.20\textwidth, clip, trim = 1cm 1.3cm 1cm 1.3cm]{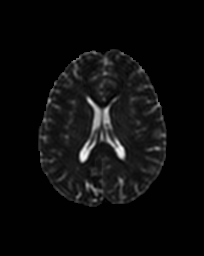} \\
   \includegraphics[width = 0.20\textwidth, clip, trim = 1cm 1.3cm 1cm 1.3cm]{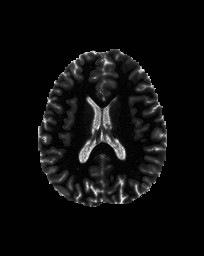} &
   \includegraphics[width = 0.20\textwidth, clip, trim = 1cm 1.3cm 1cm 1.3cm]{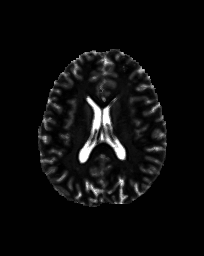} &
   \includegraphics[width = 0.20\textwidth, clip, trim = 1cm 1.3cm 1cm 1.3cm]{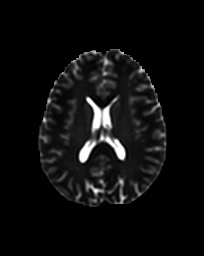} &
   \includegraphics[width = 0.20\textwidth, clip, trim = 1cm 1.3cm 1cm 1.3cm]{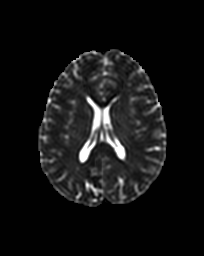} 
   \end{tabular}
\caption{\label{fig:t2synth}\textbf{T1-to-T2 Synthesis results:} 
Shown are the results of synthesis using unnormalized~(top row) and
FCM normalized images~(bottom row). The unnormalized DNN result
represents a failure of image synthesis.}
\end{figure}
}

\section{DISCUSSION AND CONCLUSION}
We have shown that: 1)~synthesis methods are substantially improved
with the addition of an intensity normalization pre-processing step, especially DNN synthesis;
2)~synthesis is robust to the choice of normalization method as we see no statistically
significant difference in the presented normalization methods.

The failure cases shown in Figs.~\ref{fig:flairsynth} and~\ref{fig:t2synth} 
results from the histogram of a particular input T1-w image being different
than the majority of T1-w images the model was trained on. In this case, the problem
histogram is compressed such that the grey matter peak was nearly aligned with
the average location of the WM peaks for all but one of the training set (where the 
outlier on the training set also has the grey matter peak in the vicinity 
of the WM peak average for the training set). 

\begin{figure}[!ht]
   \centering
   \begin{tabular}{ccc} 
   \includegraphics[width = 0.28\textwidth, clip, trim = 1.4cm 2cm 0.6cm 1.5cm]{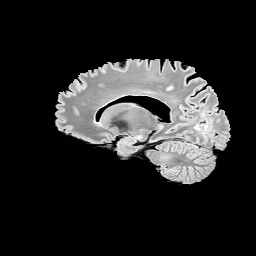} &
   \includegraphics[width = 0.28\textwidth, clip, trim = 1.4cm 2cm 0.6cm 1.5cm]{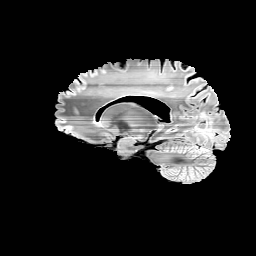} &
   \includegraphics[width = 0.28\textwidth, clip, trim = 1.4cm 2cm 0.6cm 1.5cm]{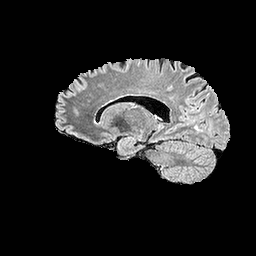}
   \end{tabular}
   \caption{\label{fig:sag} \textbf{T1-to-FLAIR DNN Synthesis} Shown from left to right are the results of DNN synthesis
          using FCM normalized images, unnormalized images, and the ground truth.}
\end{figure}

The fact that we fail to synthesize unnormalized images correctly in the best case scenario---all of our training 
and testing images came from the same cohort acquired on the same scanner with the same pulse
sequence and all of the images are of healthy patients---points to the importance of intensity normalization
as a preprocessing step in any synthesis pipeline.
While the highlighted failure case is
remarkable, the synthesized versions of the remaining images also exhibit more
subtle failure. Specifically, we see poor correspondence in intensities between slices.
That is, if you scan through the images on the plane through which the image
was synthesized (in this case axial), the result appears like a reasonable synthesis;
however, when the image is viewed in the saggital plane we see significant variation
in the intensities of neighboring slices and this variation is not observed in the 
synthesis results of normalized images (see Fig. \ref{fig:sag} for an example).
While this slice-to-slice variation is partly due to using a 2D synthesis 
method, 2D synthesis is commonly used in state-of-the-art synthesis methods\cite{huo2017adversarial,Wolterink2017,Chartsias2018}.
Since the DNN performs better across all metrics when the images are normalized, normalization 
is suggested as a pre-processing step before training or testing any sort of patch-based DNN.

\acknowledgements
This work was supported in part by the NIH/NINDS grant R01-NS070906
and by the National MS Society grant RG-1507-05243.

\bibliography{spie2019} 

\begin{thebibliography}{10}

\bibitem{Iglesias2013}
Iglesias, J.~E., Konukoglu, E., Zikic, D., Glocker, B., Leemput, K.~V., and
  Fischl, B., ``{Is synthesizing MRI contrast useful for inter-modality
  analysis?},'' in [{\em MICCAI}{\nolinebreak\hspace{0.1em}]},  (8149),
  631--638 (2013).

\bibitem{huo2017adversarial}
Huo, Y., Xu, Z., Bao, S., Assad, A., Abramson, R.~G., and Landman, B.~A.,
  ``Adversarial synthesis learning enables segmentation without target modality
  ground truth,'' in [{\em 2018 IEEE 15th International Symposium on Biomedical
  Imaging (ISBI 2018)}{\nolinebreak\hspace{0.1em}]},   1217--1220 (April 2018).

\bibitem{Roy2014}
Roy, S., Carass, A., Jog, A., Prince, J.~L., and Lee, J., ``{MR to CT
  registration of brains using image synthesis},'' {\em SPIE Medical
  Imaging}~{\bf 9034} (2014).

\bibitem{Roy2017}
Roy, S., Butman, J.~A., and Pham, D.~L., ``{Robust skull stripping using
  multiple MR image contrasts insensitive to pathology},'' {\em
  NeuroImage}~{\bf 146},  132--147 (2017).

\bibitem{Roy2013}
Roy, S., Carass, A., and Prince, J.~L., ``{Magnetic resonance image
  example-based contrast synthesis},'' {\em IEEE Transactions on Medical
  Imaging}~{\bf 32}(12),  2348--2363 (2013).

\bibitem{Jog2017}
Jog, A., Carass, A., Roy, S., Pham, D.~L., and Prince, J.~L., ``{Random forest
  regression for magnetic resonance image synthesis},'' {\em Medical Image
  Analysis}~{\bf 35},  475--488 (2017).

\bibitem{Zhao2017}
Zhao, C., Carass, A., Lee, J., Jog, A., and Prince, J.~L., ``{A supervoxel
  based random forest synthesis framework for bidirectional MR/CT synthesis},''
  {\em Simulation and Synthesis in Medical Imaging (SASHIMI)}~{\bf 10557
  LNCS}(1),  33--40 (2017).

\bibitem{Lee2017}
Lee, J., Carass, A., Jog, A., Zhao, C., and Prince, J.~L., ``{Multi-atlas-based
  CT synthesis from conventional MRI with patch-based refinement for MRI-based
  radiotherapy planning},'' in [{\em SPIE Medical Imaging~(SPIE-MI
  2017)}{\nolinebreak\hspace{0.1em}]},   {\bf 10133} (2017).

\bibitem{Cardoso2015}
Cardoso, M.~J., Sudre, C.~H., Modat, M., and Ourselin, S., ``{Template-based
  multimodal joint generative model of brain data},'' in [{\em Information
  Processing in Medical Imaging}{\nolinebreak\hspace{0.1em}]},   {\bf 9123},
  17--29 (2015).

\bibitem{Wolterink2017}
Wolterink, J.~M., Dinkla, A.~M., Savenije, M.~H., Seevinck, P.~R., van~den
  Berg, C.~A., and I{\v{s}}gum, I., ``{Deep MR to CT synthesis using unpaired
  data},'' in [{\em Lecture Notes in Computer
  Science}{\nolinebreak\hspace{0.1em}]},   {\bf 10557 LNCS},  14--23 (2017).

\bibitem{Chartsias2018}
Chartsias, A., Joyce, T., Giuffrida, M.~V., and Tsaftaris, S.~A., ``{Multimodal
  MR Synthesis via Modality-Invariant Latent Representation},'' {\em IEEE
  Transactions on Medical Imaging}~{\bf 37}(3),  803--814 (2018).

\bibitem{Nyul2000}
Ny{\'{u}}l, L.~G., Udupa, J.~K., and Zhang, X., ``{New Variants of a Method of
  MRI Scale Standardization},'' {\em IEEE Transactions on Medical Imaging}~{\bf
  19}(2),  143--150 (2000).

\bibitem{Shah2011}
Shah, M., Xiao, Y., Subbanna, N., Francis, S., Arnold, D.~L., Collins, D.~L.,
  and Arbel, T., ``{Evaluating intensity normalization on MRIs of human brain
  with multiple sclerosis},'' {\em Medical Image Analysis}~{\bf 15}(2),
  267--282 (2011).

\bibitem{Shinohara2014}
Shinohara, R.~T., Sweeney, E.~M., Goldsmith, J., Shiee, N., Mateen, F.~J.,
  Calabresi, P.~A., Jarso, S., Pham, D.~L., Reich, D.~S., and Crainiceanu,
  C.~M., ``{Statistical normalization techniques for magnetic resonance
  imaging},'' {\em NeuroImage: Clinical}~{\bf 6},  9--19 (2014).

\bibitem{Fortin2016}
Fortin, J.~P., Sweeney, E.~M., Muschelli, J., Crainiceanu, C.~M., and
  Shinohara, R.~T., ``{Removing inter-subject technical variability in magnetic
  resonance imaging studies},'' {\em NeuroImage}~{\bf 132},  198--212 (2016).

\bibitem{Leek2007}
Leek, J.~T. and Storey, J.~D., ``{Capturing heterogeneity in gene expression
  studies by surrogate variable analysis},'' {\em PLoS Genetics}~{\bf 3}(9),
  1724--1735 (2007).

\bibitem{Avants2008}
Avants, B.~B., Epstein, C.~L., Grossman, M., and Gee, J.~C., ``{Symmetric
  diffeomorphic image registration with cross-correlation: Evaluating automated
  labeling of elderly and neurodegenerative brain},'' {\em Medical Image
  Analysis}~{\bf 12}(1),  26--41 (2008).

\bibitem{Ronneberger2015}
Ronneberger, O., Fischer, P., and Brox, T., ``{U-Net: Convolutional Networks
  for Biomedical Image Segmentation},'' in [{\em
  MICCAI~2015}{\nolinebreak\hspace{0.1em}]},  {\em Lecture Notes in Computer
  Science} {\bf 9351},  234--241, Springer Berlin Heidelberg (2015).

\bibitem{Zhao2017a}
Zhao, C., Carass, A., Lee, J., He, Y., and Prince, J.~L., ``{Whole brain
  segmentation and labeling from CT using synthetic MR images},'' {\em Machine
  Learning in Medical Imaging (MLMI)}~{\bf 10541},  291--298 (2017).

\bibitem{Wang2004}
Wang, Z., Bovik, A.~C., Sheikh, H.~R., and Simoncelli, E.~P., ``{Image quality
  assessment: From error visibility to structural similarity},'' {\em IEEE
  Transactions on Image Processing}~{\bf 13}(4),  600--612 (2004).

\bibitem{Landman2011}
Landman, B.~A., Huang, A.~J., Gifford, A., Vikram, D.~S., Lim, I. A.~L.,
  Farrell, J.~A., Bogovic, J.~A., Hua, J., Chen, M., Jarso, S., Smith, S.~A.,
  Joel, S., Mori, S., Pekar, J.~J., Barker, P.~B., Prince, J.~L., and van Zijl,
  P.~C., ``{Multi-parametric neuroimaging reproducibility: A 3-T resource
  study},'' {\em NeuroImage}~{\bf 54}(4),  2854--2866 (2011).

\bibitem{Tustison2010}
Tustison, N.~J., Avants, B.~B., Cook, P.~A., Zheng, Y., Egan, A., Yushkevich,
  P.~A., and Gee, J.~C., ``{N4ITK: Improved N3 bias correction},'' {\em IEEE
  Transactions on Medical Imaging}~{\bf 29}(6),  1310--1320 (2010).

\bibitem{avants2009advanced}
Avants, B.~B., Tustison, N., and Song, G., ``Advanced normalization tools
  ({ANTS}),'' {\em Insight j}~{\bf 2},  1--35 (2009).

\bibitem{Iglesias2011}
Iglesias, J.~E., Liu, C.~Y., Thompson, P.~M., and Tu, Z., ``{Robust brain
  extraction across datasets and comparison with publicly available methods},''
  {\em IEEE Transactions on Medical Imaging}~{\bf 30}(9),  1617--1634 (2011).

\end{thebibliography}
\bibliographystyle{spiebib} 

\end{document}